\pdfoutput=1

\documentclass[11pt]{article}
\usepackage[final]{acl}
\usepackage{times}  
\usepackage{helvet} 
\usepackage{courier}
\usepackage{graphicx}
\usepackage{caption}
\usepackage{algorithm}
\usepackage{algorithmic}
\usepackage{newfloat}
\usepackage{listings}
\usepackage{microtype}
\usepackage{booktabs}
\usepackage{amsmath}
\usepackage{breqn}
\usepackage{multirow}

\DeclareCaptionStyle{ruled}{labelfont=normalfont,labelsep=colon,strut=off} 
\floatstyle{ruled}
\newfloat{listing}{tb}{lst}{}
\floatname{listing}{Listing}

\usepackage[most]{tcolorbox}
\newtcblisting{prompt}{
   enhanced,
    colback=white,
    colframe=gray!75,
    listing only,
    listing options={
        basicstyle=\ttfamily\small,
        breaklines=true,
        showstringspaces=false,
        columns=fullflexible,
        frame=none,
        numbers=none,
        escapechar={|},
        breakindent=0pt
    },
    left=1mm,
    right=1mm,
    top=1mm,
    bottom=1mm,
    arc=1mm,
    boxrule=0.5pt
}

\DeclareMathOperator*{\argmax}{\arg\!\max}
\DeclareMathOperator*{\argmin}{\arg\!\min}

\title{K-Edit: Language Model Editing with Contextual Knowledge Awareness}

\author{Anil Ramakrishna\textsuperscript{1}, Yixin Wan\textsuperscript{2}, Xiaomeng Jin\textsuperscript{3}, Kai-Wei Chang\textsuperscript{1,2}, Zhiqi Bu\textsuperscript{1}, \\ \textbf{Bhanukiran Vinzamuri}\textsuperscript{1}, \textbf{Volkan Cevher}\textsuperscript{1,4}, \textbf{Mingyi Hong}\textsuperscript{1,5}, \textbf{Rahul Gupta}\textsuperscript{1} 
}

\author{
    Elan Markowitz,\textsuperscript{\rm 1, 2} 
    Anil Ramakrishna,\textsuperscript{\rm 1}
    Ninareh Mehrabi,\textsuperscript{\rm 1}
    Charith Peris,\textsuperscript{\rm 1}\\
    \textbf{Rahul Gupta},\textsuperscript{\rm 1}
    \textbf{Kai-Wei Chang},\textsuperscript{\rm 1, 3}
    \textbf{Aram Galstyan}\textsuperscript{\rm 1}
    \\
\textsuperscript{1}Amazon AGI,  \\ \textsuperscript{2}University of Southern California - Information Sciences Institute, \\  \textsuperscript{3}University of California - Los Angeles \\
\texttt{esmarkow@usc.edu}
}

\begin{document}
\maketitle
\begin{abstract}
As the world changes, we need to be able to update our models and correct false information without costly retraining. Knowledge-based model editing enables precise modifications to the weights of large language models in order to modify the information encoded within. Recent approaches have seen success in enabling recall of edited information for thousands of edits at once. However, these approaches fail to produce edits that account for associated contextual information. We present K-Edit, an effective approach to generating contextually consistent knowledge edits. By using knowledge graphs, which maintain contextual consistency when an edge is edited, we are able to generate additional \textit{contextual edits} that ensure consistency of related information in the language model. Our experiments demonstrate significant improvements in multi-hop question answering while maintaining the general effectiveness and scalability of model edits. 
\end{abstract}

\section{Introduction}

\begin{figure*}[h!]
    \centering
    \includegraphics[width=\textwidth]{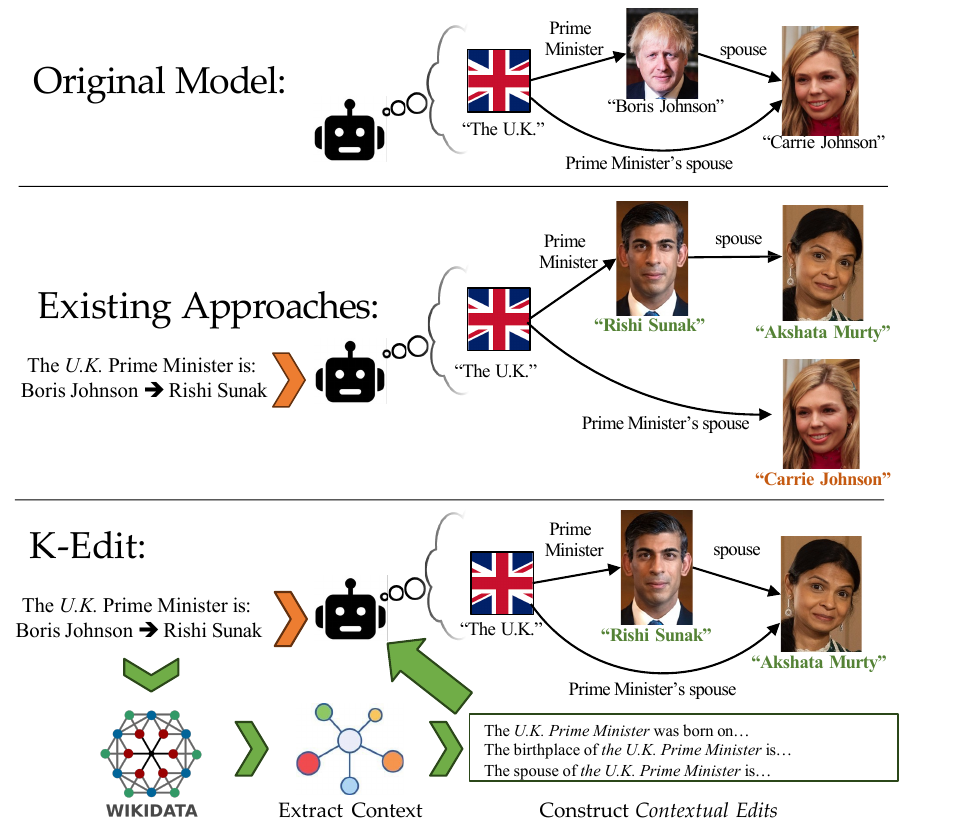}
    \caption{Example of how K-Edit improves model editing. Whereas existing methods edit the model so that it recalls the correct updated information, that new information is not associated with contextual information. The result is that the LLM is left with an inconsistent world model in which ``The Prime Minister is Rishi Sunak'', and ``Rishi Sunak is married to Akshata Murty'', but ``The Prime Minister's spouse is Carrie Johnson''. K-Edit fixes this by querying a knowledge graph (such as Wikidata) for contextual information, and then turning that information into \textit{contextual edits}, and applying it to the LLM to improve multi-hop reasoning and consistency.}
    \label{fig:main}
\end{figure*}
The ability to pinpoint and edit knowledge within the parameters of a Large Language Model (LLM) is a powerful tool for maintaining alignment with an ever-changing world. Existing approaches to direct model editing make shallow edits that fail to integrate with contextual knowledge the model possesses \citep{Zhong2023MQuAKEAK}. This work proposes and explores a simple solution to better enhance knowledge reasoning over edits by including contextual and associated knowledge in the editing process. We use the contextual consistency of knowledge graphs when edited to improve the consistency of language models under direct model editing. 

Direct Model Editing (or Knowledge-based Model Editing) aims to change the implicit knowledge in an LLM's weights with no other changes to how the model is used \citep{Wang2023KnowledgeEF, Mazzia2023ASO, Yao2023EditingLL}. Prior methods have focused on the ability to edit and recall a new fact without affecting neighboring facts or general generation ability \citep{ftZhu2020ModifyingMI, mendMitchell2021FastME, romeMeng2022LocatingAE, memitMeng2022MassEditingMI}. Recently, however, researchers have begun to question the effectiveness of these edits through benchmarks evaluating the ability of models to use these edits for multi-hop reasoning \citep{Zhong2023MQuAKEAK}. For example, if we edit the British Prime Minister to be \textit{Rishi Sunak} instead of \textit{Boris Johnson}, we would expect the model to be able to answer \textit{Who is the British Prime Minister's wife?} with the correct answer \textit{Akshata Murty}. Existing direct model editing techniques largely fail in this regard. We address this issue by utilizing the relational information in knowledge graphs. 

In this work, we propose K-Edit, a simple, yet effective, approach of adding more contextual knowledge into the editing process. In contrast to text (and thus LLMs), knowledge graphs are very simple to edit. Knowledge graphs are graph databases encoding the relations between real-world entities. Since the structure of knowledge is explicit in a knowledge graph, editing a single edge immediately updates connected associations as well. We propose a method that incorporates these automatic associations into the editing process. After an edit is made, we extract related contextual information from a knowledge graph. We then construct \textit{contextual edits} that ensure the context is ingrained in addition to the initial edit. Figure \ref{fig:main} shows the advantage of this approach.

Our results show that this method significantly helps with the primary goal, as measured by significant improvement on multi-hop reasoning over edited information. We even show that such edits improve multi-hop reasoning even when the required reasoning path does not directly appear in the context. Our results also indicate that there is no cost to quality as measured by other knowledge editing metrics, and that thousands of such edits can be applied without degradation compared to baselines. 

Our contributions are

\begin{enumerate}
    \item K-Edit, a new direct model editing algorithm for improving contextual and associated knowledge awareness through incorporation of knowledge graph data. 
    \item Experimental evaluation that shows improved multi-hop reasoning with K-Edit and its ability to generalize.
    \item Additional experiments and ablations showing no degradation to other knowledge editing metrics and the ability to apply thousands of such edits at a time. 
\end{enumerate}

\section{Background}

\subsection{Knowledge Graphs}

Knowledge graphs encode real world entities and the relations between them. The most prominent knowledge graphs contain billions of edges covering a vast array of subjects, such as Wikidata \citep{Vrandei2014WikidataAF} and DBPedia \citep{Lehmann2015DBpediaA}. We define a knowledge graph $\mathcal{K} = (\mathcal{E},\mathcal{R},\mathcal{T})$ where $\mathcal{E}$ is the set of entities, $\mathcal{R}$ is the set of relations, and $\mathcal{T}$ is the set of edges $(s,r,o)\in \mathcal{E}\times \mathcal{R} \times \mathcal{E}$ that represent true relationship in the knowledge base between subject $s$ and $o$. For example, if $(\textit{Blank Space, performer, Taylor Swift}) \in \mathcal{T}$, then the statement ``The performer of Blank Space is Taylor Swift'' is considered true. One feature of knowledge graphs is that if an edge is updated, all multi-hop relations associated with that edge will be correct without any additional explicit updates. 

\subsection{Knowledge-based Model Editing}

We define some background notation and concepts for knowledge-based model editing. Here we deal with edits of the form, $(s,r,o\rightarrow o^*)$, meaning we want to edit subject $s$ such that the relationship $r$ now refers to object $o*$ instead of $o$. We define a text prompt $t_r(s)$ for the edit (also referred to as a cloze statement) in which $o^*$ is the desired output, e.g. $t_{\text{prime minister}}(\text{\small{United Kingdom}})$ = ``The Prime Minister of \{\textit{the United Kingdom}\} is''. In this example \textit{the United Kingdom} is the entity for which we want to modify the \textit{Prime Minister} association. In general, knowledge-based model editing approaches seek to modify the model parameters $\theta$ such that the prompt $t_r(s)$ will lead to the generation of text $o^*$. 

\subsection{Batch Editing}

In order for our approach, K-Edit, to succeed, we rely on the ability to make edits in a batch. Our approach works with any such algorithm. For our experiments, we use MEMIT, a state-of-the-art batch editing algorithm that can scale to thousands of edits at once \citep{memitMeng2022MassEditingMI}. MEMIT is the best option to use with K-Edit, and the only option that can scale to large numbers of edits.

MEMIT has three key properties: (1) it computes the desired model parameter changes for thousands of edits independently, (2) it precisely targets associations of the subjects to be edited, and (3) it can apply all edits together without significant model degradation. Further background details on how MEMIT works can be found in Appendix \ref{sec:app_memit} as well as in their paper. 

\section{K-Edit Algorithm}

K-Edit leverages knowledge graphs to improve the quality of model edits by incorporating contextual information. The algorithm consists of three main steps that work together to create more comprehensive knowledge updates.

Existing approaches modify representations of the subject $s$ such that the model generates the desired association $o^*$ for relationship $r$. However, generating the associated text is not the same as internalizing the association. K-Edit uses external knowledge bases to improve the quality of model edits such that the new subject representation is integrated with connected contextual associations.

K-Edit works according to the following steps. (1) initial edits are applied the model using a batch editing approach (e.g. MEMIT). (2) Contextual knowledge is extracted from a knowledge graph around the edit and converted to \textit{contextual edits}. (3) The contextual edits are applied using the same batch editing approach from step 1. This can be repeated for deeper depths of multi-hop contextual knowledge. Algorithm \ref{alg:k_edit} shows the high level algorithm for implementing K-Edit. 


\begin{algorithm}[tb]

    \caption{K-Edit Algorithm}
    \begin{algorithmic}[1]
    \label{alg:k_edit}
        \REQUIRE Model $\theta_0$, Batch of initial edits $B$, \\KG $\mathcal{K}=(\mathcal{E},\mathcal{R},\mathcal{T})$, context depth $k$
        \STATE $\theta_1 \leftarrow \text{BatchEdit}(\theta_0, B)$
        \FOR{$i \in [2, ..., k]$}
            \STATE $C \leftarrow \text{GetContextualEdits}(B, \mathcal{K}, i)$ \hfill\textcolor{blue}{\textit{\# edits of depth i}}
            \STATE $\theta_{i} \leftarrow \text{BatchEdit}(\theta_{i-1}, C)$
        \ENDFOR
        \RETURN $\theta_k$ \hfill\textcolor{blue}{\textit{\# Returns the final model parameters}}
    \end{algorithmic}
\end{algorithm}

\subsection{Creating Contextual Edits}


\begin{algorithm}[tb]
    \caption{GetContextualEdits}
    \begin{algorithmic}[1]
    \label{alg:contextual_edits}
        \REQUIRE Batch of initial edits $B$, KG $\mathcal{K}=(\mathcal{E},\mathcal{R},\mathcal{T})$,\\ current context depth $d$
        \STATE $C_1 \gets B$
        \FOR{$i$ in $[2,...,d]$}
            \STATE $C_i \gets \emptyset$ \hfill\textcolor{blue}{\textit{\# Contextual edits of depth i}}
            \FORALL{$(s,r_1,...,r_{i-1}\rightarrow o_{i-1}^*) \in C_{i-1}$}
                \STATE \textcolor{blue}{\textit{\# Next-hops based on previous depth's target object}}
                \STATE $N \gets \{(s_i, r_i, o_i) \in \mathcal{T} \, | \, s_i = o_{i-1}^*\}$
                \FORALL{$(o_{i-1}^*, r_i, o_i^*) \in N$}
                    \STATE \textcolor{blue}{\textit{\# Construct new edits using the next-hop edges}}
                    \STATE $e \gets (s,r_1,...,r_{i-1},r_i\rightarrow o_i^*)$
                    \STATE $C_i \gets C_i \cup \{e\}$ 
                \ENDFOR
            \ENDFOR
        \ENDFOR
        \STATE $O \gets \emptyset$
        \FORALL{$(s,r_1,...,r_d \rightarrow o_d^*) \in C_d$} 
            \STATE \textcolor{blue}{\textit{\# Create text prompts for each edit of depth d}}
            \STATE $O \gets O \cup \{t_{r_d}(r_{d-1}(...(r_1(s))) \rightarrow o_2\}$ 
        \ENDFOR
        \RETURN $O$\hfill\textcolor{blue}{\textit{\# Returns all contextual edits of depth d}}
    \end{algorithmic}
\end{algorithm}

Contextual edits are model editing prompts that are constructed to not just reinforce a simple edited fact, but also to ensure the fact is linked with connected facts (Examples of connected facts in figure \ref{fig:main}). With this goal in mind, we design contextual edits  that both require the model to recall the edited information on its own and require the model to follow the multi-hop relationships. The following section describes the construction of such prompts. 

For a given edit $(s,r_1,o_1\rightarrow o_1^*)$, we compute $k$-hop edits by sampling relationships from the neighborhood of $o_1^*$, $\mathcal{N}(o_1^*)$. We then combine these edges with the original edit to get multi-hop edges of the form $(s, r_1, o_1^*), (o_1^*, r_2, o_2)$ e.g. (United Kingdom, Prime Minister, \textit{Rishi Sunak}), (\textit{Rishi Sunak}, spouse, Akshata Murty). Simply applying the edit (Rishi Sunak, spouse, Akshata Murty$\rightarrow$Akshata Murty) would not work as the model already knows this information $(o_1^*, r_2, o_2\rightarrow o_2)$. Instead, we construct the edit statement in a way that forces the model to incorporate both the edited edge and associated edge together. Specifically, we construct an edit of the format $t_{r_2}(r_1(s))\rightarrow o_2$ e.g. ``The spouse of \{\textit{the United Kingdom Prime Minister}\} is$\rightarrow$Akshata Murty''. Here, $r_1(s)$ is a text template that maps to the object referred to from $s$ on relation $r_1$ and presents it in text as a subject-relation clause. We can see that this edit does not include $o_1^*$ explicitly. As a result, the model must rely on the previously updated association $(s,r_1,o_1\rightarrow o_1^*)$ in order to complete the prompt. 

We also note that we treat \textit{the United Kingdom Prime Minister} as the subject to edit which means the [Minister] token rather than [Kingdom] token is the one that is the target of the update. We do this in accordance with the findings of \citet{Geva2023DissectingRO} that the final token of a subject aggregates the information and relationships from preceding tokens. This process allows the model to learn the 2nd hop association in the context of the previous edit as opposed to memorizing all 2-hop associations for the subject token. We note that this setup is coincidentally similar to the approach used to analyze models in \citet{Yang2024DoLL}. It is encouraging that our approach to edit multi-hop associations corresponds to the way they discovered models perform latent multi-hop reasoning in the first place. Algorithm \ref{alg:contextual_edits} presents the detailed description for constructing contextual edits of any hop depth.

\subsection{Scaling Contextual Edits}

One of the challenges with generating multi-hop contextual edits is that the number of templates for editing rapidly explodes. For instance, with 50 relationship types, there are potentially 2,500 different templates required for 2-hop contextual edits (though most of those will not exist in practice). This could be solved naively by generating the templates with an LLM but that could still get burdensome for larger KGs or hop depths and could introduce errors. Instead, our approach employs a linguistic mechanism that enables the scalable creation of contextual edits \textit{for any hop length}.

To use this method all that is needed is two text templates for each relation type $r$. We require one template $t_r(s)$ that creates a cloze statement for each relation $r$. This is a fill-in-the-blank style prompt. For example, the template ``\textit{The Prime Minister of \{s\} is}" prompts the model to complete the association. Additionally we need one more template, $r(s)$ that creates a subject-relation clause that semantically refers to the corresponding object e.g. "\textit{the spouse of \{s\}}". With these mappings we create a single cloze statement template for a multi-hop edit, using the template $t_{r_1}(r_2(...(r_k(s)))$. This means we only need to create $2\times |\mathcal{R}|$ templates rather than $|\mathcal{R}|^k$. Note that their is no dependence on depth $k$. Algorithm 2 presents the method for constructing these contextual edits. These more simple templates enable the easy use of LLMs for further generation (Appendix \ref{sec:llm_for_template_creation}).

\begin{table*}
\centering
\begin{minipage}[t]{0.49\textwidth}
    \centering
    \begin{tabular}{@{}lcc@{}}
        \toprule
         & MQuAKE & MultiCounterfact  \\ \midrule
        \#examples & 3,000 & 10,000 \\
        \#relations & 36 & 34 \\ \bottomrule
    \end{tabular}
    \caption{Dataset statistics.}
    \label{tab:data_stats}
\end{minipage}
\hfill
\begin{minipage}[t]{0.49\textwidth}
    \centering
    \begin{tabular}{@{}l|llll@{}}
    \toprule
     & 1-edit & 2-edits & 3-edits & 4-edits \\ \midrule
    2-hops & 513 & 487 & NA & NA \\
    3-hops & 356 & 334 & 310 & NA  \\
    4-hops & 224 & 246 & 262 & 268 \\ \bottomrule
    \end{tabular}
    \caption{Breakdown by number of hops and number of edits for the MQuAKE dataset (MQuAKE-CF-3k version)}
    \label{tab:mquake_breakdown}
\end{minipage}
\end{table*}

\section{Experiments}

We run experiments on K-Edit to evaluate its ability to incorporate contextual information under edits as well as the general quality of the model edits produced. 

\subsection{Datasets}

We use two datasets, \textbf{MQuAKE} \citep{Zhong2023MQuAKEAK} and \textbf{MultiCounterFact}\citep{memitMeng2022MassEditingMI}. Dataset statistics can be found in Table \ref{tab:data_stats}.

The primary dataset, \textbf{MQuAKE}, is used for evaluating multi-hop reasoning on edits \citep{Zhong2023MQuAKEAK}. This dataset contains 3,000 multi-hop questions (2-4 hops) with each answer based on 1-4 counterfactual edits. For example, the question ``What is the headquarters location for the broadcaster of True Detective?'' is paired with two edits: (True Detective, broadcaster, HBO $\rightarrow$ ABC) and (ABC, headquarters location, New York City $\rightarrow$ Los Angeles). These edits are not real in the world, but are plausible counterfactual edits that could be applied. This example was a two-hop question with two edits, however, the number of edits does not need to be the same as the number of hops. The exact breakdown can be found in Table \ref{tab:mquake_breakdown}. 

We also use a subset of \textbf{MultiCounterFact} dataset \citep{memitMeng2022MassEditingMI} derived from CounterFact \citep{romeMeng2022LocatingAE}. This dataset contains 30,000 counterfactual edits and is used to evaluate the strength of an edit, whether unaffected neighboring relations remain unchanged, and the generation ability of the model post editing. The MultiCounterFact modification ensures there are no direct conflicts between edits in the dataset (i.e. no duplicate edits for a given subject and relation). We use a subset of 10,000 questions for our experiments. 

\begin{table*}[t]
\centering
\begin{tabular}{@{}llcc@{}}
\toprule
Model & Algorithm & \multicolumn{1}{l}{Multi-Hop} & \multicolumn{1}{l}{Multi-Hop CoT} \\ \midrule
\multirow{6}{*}{\textbf{GPT-J-6B}} & Base$\dagger$ & 43.4 & 42.1 \\ \cmidrule(l){2-4} 
 & FT-W$\dagger$ & 1.6 & 1.9 \\
 & MEND$\dagger$ & 9.2 & 11.5 \\
 & ROME$\dagger$ & 7.6 & \textbf{18.1} \\
 & MEMIT$\dagger$ & 8.1 & 12.3 \\
 & \textbf{K-Edit} & \textbf{14.5 (0.64)} & 17.0 (0.63) \\ \midrule
\multirow{2}{*}{\textbf{Falcon-7b}} & Base & 38.1 & 43.3 \\ \cmidrule(l){2-4} 
& MEMIT & 5.87 & 12.57 \\
 & \textbf{K-Edit} & \textbf{11.1 (0.57)} & \textbf{18.43 (0.71)} \\ \bottomrule
\end{tabular}
\caption{Results on MQuAKE multi-hop question answering and reasoning dataset over edited information. Results show that K-Edit significantly improves Multi-Hop QA and is similar to ROME on chain-of-thought question answering. Base gives the results of the unedited model on the original answers. Parenthesis show standard error of the mean estimate. $\dagger$ indicates results from \citet{memitMeng2022MassEditingMI}}
\label{tab:mquake}
\end{table*}

\begin{table*}[t]
\centering
\begin{tabular}{@{}llcc@{}}
\toprule
Model & Algorithm & \multicolumn{1}{l}{Multi-Hop} & \multicolumn{1}{l}{Multi-Hop CoT} \\ \midrule
 & MEMIT$\dagger$ & 8.1 & 12.3 \\
 \textbf{GPT-J-6B} & K-Edit (Generalize) & 11.5 (0.58) & 14.8 (0.65) \\
 & \textbf{K-Edit} & \textbf{14.5 (0.64)} & \textbf{17.0 (0.63)} \\ \midrule
 & MEMIT & 5.87 & 12.57 \\
\textbf{Falcon-7b} & K-Edit (Generalize) & 8.2 (0.50) & 16.3 (0.68) \\
 & \textbf{K-Edit} & \textbf{11.1 (0.57)} & \textbf{18.4 (0.71)} \\ \bottomrule
\end{tabular}
\caption{Generalization experiment in which the multi-hop edge being evaluated is removed from the contextual edits. This shows that some of the K-Edit improvement stems from improved representations rather than more broad memorization of associations. Results are for MQuAKE dataset. Parenthesis show standard error of the mean estimate. $\dagger$ indicates results from \citet{memitMeng2022MassEditingMI}}
\label{tab:generalize}
\end{table*}

\subsection{Baselines}

We compare against a number of alternative direct model editing baselines. 

\textbf{FT-W} is a finetuning approach with weight decay to avoid altering other memories or changing unrelated model behavior \citep{ftZhu2020ModifyingMI}.

\textbf{MEND} uses a hypernetwork meta-trained to modify fine-tuning gradients so that they produce higher quality edits \citep{mendMitchell2021FastME}.

\textbf{ROME} uses counterfactual tracing to locate what layer of the model contains the information to be edited before updating the MLP component in that layer \citep{romeMeng2022LocatingAE}.

\textbf{MEMIT} (presented above) uses a similar objective to ROME but modified to enable editing large sets of facts at once, which none of the other baselines are able to do well \citep{memitMeng2022MassEditingMI}. We consider this to be the main baseline as it is also the editing mechanism we use in our K-Edit implementation.

While there are a number of knowledge editing approaches based on retrieval mechanisms, these methods fundamentally alter the nature of the model. Our focus is on direct model editing approaches. This is discussed further in the related works section. 

\section{Results and Analysis}

Our results show that K-Edit improves multi-hop contextual reasoning over edited information without sacrificing in terms of scalability or individual edit performance. 

Table \ref{tab:mquake} shows our main results for GPT-J and Falcon-7b on the MQuAKE dataset. We see that the model's multi-hop answering ability vastly exceeds the other baselines with a relative improvement of 57.6\% over the best baseline MEND (14.5\% versus 9.2\%) and close to doubling the performance of ROME and MEMIT (+90.8\% and +79.0\% respectively). Under Chain-of-Thought prompting, K-Edit significantly outperforms MEMIT while slightly underperforming ROME. We note that the MQuAKE results are from applying the edits for one question at a time. ROME has noted weaknesses when applying edits in bulk as will be discussed later. 

We also compare K-Edit to MEMIT for Falcon-7b and find a even greater improvement than for GPT-J with the multi-hop accuracy nearly doubling. 

While K-Edit significantly improves multi-hop accuracy, we note that there is still room for improvement. The GPT-J Base model gets 43.4\% accuracy on the unedited answers which sets a rough upper bound for where we could expect to see results.

\begin{table}[t]
\centering
\begin{tabular}{@{}lccc@{}}
\toprule
Algorithm & \multicolumn{3}{c}{Num Hops} \\ \cmidrule(l){2-4} 
& 2 & 3 & 4 \\ \midrule
MEMIT$\dagger$ & 18.2 & 4.8 & 10.1\\
  K-Edit (Generalize) & +1.9 & +5.2 & +4.2 \\
 K-Edit (Memorize) & +6.0 & +0.3 & +0.4 \\ \bottomrule
\end{tabular}
\caption{Breakdown by number of hops showing the incremental accuracy gain over MEMIT due to generalization and memorization. The gain from generalization is spread across  all the question types. While as expected the gain from memorization is concentrated on the two hop questions. Results are for GPT-J-6B on the MQuAKE dataset using CoT reasoning.}
\label{tab:generalize_hops}
\end{table}

\begin{table*}[t]
\centering
\begin{tabular}{@{}lcccccc@{}}
\toprule
Algorithm & \multicolumn{1}{l}{Score} & \multicolumn{1}{l}{Efficacy} & \multicolumn{1}{l}{Paraphrase} & \multicolumn{1}{l}{Specificity} & \multicolumn{1}{l}{Fluency} & \multicolumn{1}{l}{Consistency} \\ \midrule
FT-W@1 & 64.4 & 100.0 & 73.8 & 43.4 & 614.7 & 34.9 \\
MEND@1 & 62.8 & 97.2 & 52.8 & 53.9 & 620.6 & 32.3 \\
ROME@1 & \textbf{91.9} & \textbf{99.9} & \textbf{99.5} & 79.5 & 619.9 & \textbf{42.5} \\
MEMIT@1 & 91.6 & \textbf{99.8} & 95.2 & \textbf{81.7} & 621.5 & 41.5 \\
\midrule
\textbf{K-Edit@1} & 90.1 & \textbf{100.0 (0.0)} & 91.6 (0.85) & \textbf{80.8 (0.98)} & \textbf{622.2 (0.53)} & 40.7 (0.44) \\
\midrule
FT-W@1,000 & 71.1 & 99.6 & 75.2 & 53.1 & 258.1 & 17.0 \\
MEND@1,000 & 48.4 & 45.2 & 46.3 & 54.9 & 527.4 & 15.4 \\
ROME@1,000 & 60.3 & 68.2 & 64.4 & 51.0 & 509.9 & 6.8 \\
MEMIT@1,000 & 90.5 & \textbf{99.7} & 93.4 & \textbf{80.6} & \textbf{622.0} & \textbf{41.4} \\ \midrule
\textbf{K-Edit@1,000} & \textbf{90.6} & \textbf{99.8 (0.14)} & \textbf{95.3 (0.56)} & \textbf{79.5 (0.81)} & 620.2 (0.55) & \textbf{41.7 (0.39)}\\
\bottomrule
\end{tabular}
\caption{Results on MultiCounterfact dataset (from MEMIT) with GPT-J. K-Edit does not degrade and in fact, may actually improve editing ability. Results@1 indicates performing a single edit at a time. @1,000 indicates performing 1,000 edits at once. Efficacy measures memorization of the edit. Paraphrase measures robustness to paraphrases. Specificity measures whether other similar facts remain unchanged. Fluency measures entropy of generated text. Consistency measures the TF-IDF similarity between generated text about the subject and a reference text about the new object. Formal definitions are in the appendix.}
\label{tab:mcf}
\end{table*}

\begin{table}[t]
    \centering
    \begin{tabular}{lcc}
        \toprule
        Algorithm & Multi-Hop & Multi-Hop CoT \\
        \midrule
        MEMIT & 8.1 & 12.3\\
        K-Edit-Parallel & 12.6 (0.61) & 12.7(0.61) \\
        \textbf{K-Edit} & \textbf{14.5 (0.64)} & \textbf{17.0 (0.63)}\\
        \bottomrule
    \end{tabular}
    \caption{Results showing the necessity of sequential updates in K-Edit.}
    \label{tab:sequential_vs_parallel}
\end{table}

\begin{table*}[t]
\hspace{0.4cm}
\begin{minipage}[t]{0.45\textwidth}
\centering
\begin{tabular}{|c|lll|l|}
    \toprule
    \multicolumn{5}{c}{Multi-Hop Accuracy}\\ \midrule
    Edits\textbackslash Hops & 2 & 3 & 4 & All \\ \midrule
    1 & 30.6 & 7.0 & 28.1 & 22.4 \\
    2 & 15.6 & 9.3 & 16.67 & 13.9 \\
    3 & NA & 2.3 & 5.0 & 3.5 \\
    4 & NA & NA & 8.2 & 8.2 \\ \midrule
    All & 23.3 & 6.3 & 13.9 &  \\ \bottomrule
\end{tabular}
\end{minipage}
\centering
\hspace{0.2cm}
\begin{minipage}[t]{0.45\textwidth}
\begin{tabular}{|c|lll|l|}
    \toprule
    \multicolumn{5}{c}{Multi-Hop CoT Accuracy}\\ \midrule
    Edits\textbackslash Hops & 2 & 3 & 4 & All \\ \midrule
    1 & 31.0 & 5.6 & 20.5 & 20.6 \\
    2 & 20.9 & 7.8 & 21.5 & 17.0 \\
    3 & NA & 18.4 & 12.2 & 15.6 \\
    4 & NA & NA & 6.0 & 6.0 \\ \midrule
    5 & 26.1 & 10.3 & 14.7 &  \\ \bottomrule
\end{tabular}
\end{minipage}
\caption{GPT-J multi-Hop accuracy on MQuAKE by number of hops and number of edits.}
\end{table*}


\subsection{Generalization}

One question that may arise is whether the benefit of K-Edit is simply from the multi-hop questions being covered by the contextual edits. Specifically, are the observed gains due to brute force inclusion of related information or is there an aspect of K-Edit that improves the edited representations in a more generalized way? To address this question, we run additional ablation experiments in which the question edges from MQuAKE are filtered out of the contextual edits. We present these experiments in Table \ref{tab:generalize}. We see that while performance does decrease (as expected), it is still significantly better than the original MEMIT performance. Specifically, we observe that roughly half of the improvement from K-Edit generalizes to edges not covered by contextual edits. This is a somewhat surprising result. Including contextual edits on tangentially related edges improves the quality of original edit such that other multi-hop information is easier for the model to retrieve. The implication of this is that the quality of K-Edit is not solely limited to the relation types present in the knowledge graph.

When we break down these gains by the number of hops, we see that the gains from generalization are spread across the dataset, with gains on 2, 3, and 4-hop questions. Table \ref{tab:generalize_hops} shows the incremental gain from generalization and from memorization on GPT-J using CoT reasoning. We see that the incremental gain from memorized associations are concentrated on the 2-hop questions. This is expected since we only use 2-hop contextual edits in our experiments. This is also evidence that K-Edit could perform even better if contextual edits of greater depth were included. 

When we filter the analysis for questions with only a single edited hop, we find only a 2.8\% accuracy difference performance between full K-Edit and the generalization ablation, and absolutely no difference if employing CoT reasoning. This makes sense since questions with multi-edits would be especially hard to generalize. Generalization relies on using some degree of the LLM's implicit knowledge. When that knowledge is also edited, it would be much harder to associate the correct contextual information.

\subsection{Edit Quality Remains High for Standard Metrics and Larger Scales}

K-Edit's edit quality remains high when evaluated under metrics other than multi-hop accuracy (see Table \ref{tab:mcf}). These experiments show the results of running editing algorithms when only a single edit is applied at a time (@1) or when batches of 1,000 edits are applied at a time (@1,000). We see that K-Edit is at or near state-of-the-art for all metrics for both @1 and @1,000. While ROME slightly outperforms K-Edit in @1 results, it dramatically underperforms in results @1,000. The overall score (harmonic mean of efficacy, paraphrase, and specificity) for ROME@1,000 is 60.3 relative to K-Edit's 90.6. 

Compared to MEMIT we see highly comparable scores across the board. We note that while K-Edit does introduce additional edits in the form of \textit{contextual edits}, these edits are primarily there to reinforce the associated information of the initial edit. As a result, there is no noticeable impact to scalability. Though further research is needed to understand the full scalability limits of K-Edit. 

Unfortunately, while the edits in the MultiCounterFact dataset do not conflict with each other directly, the contextual edits of K-Edit might conflict with the dataset as we treat the knowledge graph as a source of truth when applying the contextual edits. For example, if one question includes the edit (LA Lakers, owner, Jeanie Buss$\rightarrow$ Bill Gates), K-Edit would include the contextual edit for ``The owner of the LA Lakers was born in$\rightarrow$Seattle''. However, if another question included the edit (Bill Gates, place of birth, Seattle$\rightarrow$New York), it would create a conflict. This could result in lower Specificity, Efficacy, and Generalization for K-Edit@1,000. However, this is not common and the affect is not large. 

\subsection{Sequential Ordering is Important}

K-Edit applies the contextual edits \textit{after} applying the original edits. This is so that the contextual edits can refine the initial edit to incorporate the related information. We find that running K-Edit in parallel (meaning the model updates for both original edits and contextual edits are computed at the same time) leads to significant decline in performance. Table \ref{tab:sequential_vs_parallel} shows the importance of using sequential edits compared to parallel ones (14.5 to 12.6 and 17.0 to 12.7 for CoT reasoning). We theorize that when the contextual edits are not applied after the initial edits, the model may be updating the wrong associations. For instance, in our running example, if Boris Johnson is still associated as ``the United Kingdom Prime Minister'', the algorithm may be updating the spouse of Boris Johnson rather than enhancing the association of Rishi Sunak as Prime Minister. 

\subsection{Analysis by number of hops and edits}

We see an overall trend of declining performance with increasing number of hops. However, the 4-hop questions have actually better performance than the 3-hop questions. The primary reason for this is that the MQuAKE dataset contains a more limited variety of reasoning paths for 4-hops relative to 3-hops and the LLM happens to perform better with those style of questions. 

The trend for the number of edits is much clearer in that the more edits there are that affect a question, the lower the performance is. This makes sense as each edit increases the chance of the edit being improperly incorporated. There are some outliers in this trend such as for 3-hop questions, increasing the number of edits increases the CoT performance. We suspect there may also be some unexpected interactions when multiple edits are applied at once. Since, these are counterfactual edits, the surrounding context may actually imply contradictory information. For instance, if we modify (Hello, performer, Adele$\rightarrow$Taylor Swift) and change (Taylor Swift, citizenship, United States$\rightarrow$Belgium). The contextual edit for the first edit contains both "The performer of Hello was born in"$\rightarrow$"United States" which may be slightly contradictory to the "citizenship" edit. In a real world setting, the knowledge graph will change in ways that are consistent with real-world correlations. As a result, we would not need to worry about such behavior outside of rare edge cases. 

\section{Related Works}

\subsubsection{Direct Model Editing Works} These approaches include aforementioned works \cite{ftZhu2020ModifyingMI, mendMitchell2021FastME, memitMeng2022MassEditingMI, romeMeng2022LocatingAE}. Other works use a hypernetwork to predict parameter updates for a model \citet{DeCao2021EditingFK} 

\subsubsection{Retrieval Based Works} Outside of direct model editing approaches \citep{ftZhu2020ModifyingMI, romeMeng2022LocatingAE, mendMitchell2021FastME, memitMeng2022MassEditingMI, birdMa2023UntyingTR}, other approaches have sought to adapt to changing knowledge via storing external edit memories and employing various in-context learning and retrieval mechanisms. IKE and others add all edits in the LLM context \citep{ikeZheng2023CanWE, Onoe2023CanLL}; SERAC identifies if a related fact has been edited and retrieves it \citep{seracMitchell2022MemoryBasedME}; MeLLo and DeepEdit iteratively retrieve edited memories during the generation and decoding process \citep{Zhong2023MQuAKEAK, Wang2024DeepEditKE}. 
    
We note that these retrieval-based editing approaches are fundamentally different than direct model editing approaches. Direct model editing leaves the model architecture and code entirely the same with the only change being to parameter weights. This means any system developed to use the original model can use the updated model with no other changes. In contrast, retrieval-based editing systems fundamentally change how the model is used and functions. For example, MeLLo requires setting up a retrieval database, iteratively prompting the LLM, making multiple calls to the database, and prompting the LLM to create further sub-questions \cite{Zhong2023MQuAKEAK}. This fundamentally changes the characteristics of the model, increases latency, and alters model behavior for non-QA tasks. An alternate line of work aims to supplement the parametric knowledge of LLMs by augmenting the model prompt with information retrieved from an indexed corpus \cite{lewis2020retrieval} or Knowledge Graphs \cite{markowitz-etal-2024-tree}. 

\subsubsection{Additional Aspects of Model Editing} Other papers have looked into other aspects of contextual knowledge for edits such as reversibility \citep{birdMa2023UntyingTR}, multi-linguality \citep{Wang2023RetrievalaugmentedMK, Si2024MPNLM}, or other implied logical conditions given an edit \citep{easyeditWang2023EasyEditAE, rippleCohen2023EvaluatingTR, depeditLi2023EvaluatingDI}.

\section{Future Work}

Future directions include both straightforward advancements as well as further analytical directions. Simple advancements include extending K-Edit to broader and deeper contexts. This can include more relation types, increasing the number of iterations to $k>2$, and including contextual edges connected to the edit subject $s$ instead of just $o^*$. These directions could yield simple improvements and benchmark gains, especially on the MQuAKE dataset. 

Other interesting future directions include counterfactual knowledge tracing for contextual information as well as better measures of contextual knowledge in edited representations beyond output related metrics. Further, more study on enhancing the generalization effect would enable the creation of better knowledge edits. While K-Edit makes meaningful advancement in contextual integration for knowledge editing, there is still more progress to be made.

\section{Conclusion}

This paper presents a K-Edit, an algorithm for improving knowledge-based model editing to better incorporate contextual information. Our experiments show that our approach greatly improves results on multi-hop question answering tasks when edits have been applied. We further show intriguing results demonstrating the potential for better contextual generalization of edits under K-Edit. In addition, our approach scales to thousands of edits at once with no significant degradation on standard metrics. These combine to show that K-Edit is a strong and easy tool for updating and editing knowledge in LLMs while maintaining consistency of information.

\section{Limitations}

Our method and experiments have a number of limitations. Our experiments are limited to adding context from a single additional hop from the edit target $o^*$, and only in the "forward" direction. However, the evaluation contains up to four hops and also uses context from edges connected to the subject $s$ as well. This means that K-Edit could show improved results if subject context and deeper context were used. 

We are also limited by the available datasets. Currently, there are only English based datasets for evaluating knowledge editing in multi-hop settings. In addition, we do not know if the multi-hop templating would function the same way in all languages. However, this could be mitigated with falling back to LLM-based generations if needed. 

Finally, this evaluation is limited to 7 billion parameter models or less. While model editing approaches generally become more effective on more capable models, they can also become more computationally costly.

\section{Ethical Statement}

We find no major negative ethical implications of this work. Hopefully, enabling high quality model editing work will allow providers to avoid retraining models from scratch, thus avoiding costs and reducing environmental impact. 

As noted in the limitations, this work is limited to English. We hope that the construction of multi-lingual model editing and multi-hop model editing datasets will enable this work to be studied under other languages. Since many knowledge graphs, Wikidata included, are multi-lingual, we see no reason why this work could not be applied in other languages. 

\bibliography{k-edit}

\appendix

\begin{figure*}
\begin{minipage}{\linewidth}
        \begin{prompt}
Your task is to construct a new relation template of the following form. The template involves a subject {subject} and an incomplete sentence to be filled. 

<---examples-->

RELATION: plays the sport of
TEMPLATE: The sport played by {subject} is

RELATION: position held
TEMPLATE: The position held by {subject} is

RELATION: employer
TEMPLATE: The employer of {subject} is

<---end examples--->

Complete the following template

RELATION: |\{relation\}|
TEMPLATE: 
\end{prompt}
\end{minipage}
    \caption{The above prompt can be used to create relation templates of the form $t_r(s)$. These are cloze statements in which the next tokens would match the entity referred to from the subject entity \{subject\} using relation \{relation\}. To use this prompt, ``{\{relation\}}'' would be replaced with relation $r$. }
    \label{fig:prompt_for_template1}
\end{figure*}

\section{Hardware}

The hardware requirements depend more on the underlying editing software, in this case MEMIT. We run software using 4 V100-16GB GPUs and make some optimizations for using multiple-GPUs and improving memory management.


\section{Definitions for Additional Edit Metrics}

These definitions come from prior work and are used in Table \ref{tab:mcf}.

\textbf{Efficacy} is the expectation that the target generation is more likely than the original fact. 


\textbf{Paraphrase} is the expectation that the target generation is more likely than the original fact under paraphrased prompting.


\textbf{Specificity} is the expectation that neighborhood prompts remain correct after an edit. Neighborhood prompts are prompts for semantically related subjects instead of $s$.


\textbf{Fluency} is the weighted sum of the entropy of bi- and tri-gram distributions of the generated text. 

\textbf{Consistency} is computed by generating text about $s$ and checking the TF-IDF similarity with a reference Wikipedia
text about $o^*$. Specifically measured as cosine similarity between the TF-IDF vectors.

\begin{figure*}
\begin{minipage}{\linewidth}
        \begin{prompt}
Your task is to construct a new relation template of the following form. The template involves a subject {subject} and describes an entity in relation to that subject. 

<---examples-->

RELATION: plays the sport of
TEMPLATE: the sport played by {subject}

RELATION: position held
TEMPLATE: the position held by {subject}

RELATION: employer
TEMPLATE: the employer of {subject}

<---end examples--->

Complete the following template

RELATION: |{\{relation\}}|
TEMPLATE: 
\end{prompt}
\end{minipage}
    \caption{The above prompt can be used to create relation templates of the form $r(s)$. These are noun-phrase type statements in which the referenced entity object has relation \{relation\} from subject \{subject\}. To use this prompt, ``{\{relation\}}'' would be replaced with the desired relation $r$. }
    \label{fig:prompt_for_template2}
\end{figure*}

\section{MEMIT}
\label{sec:app_memit}

MEMIT \citep{memitMeng2022MassEditingMI} works as follows. MEMIT determines a small change to the $L$-th layer activation of the last token of $s$ (e.g. [\textit{Kingdom}] in the example above) such that $t_r(s)$ produces the desired output $o^*$. It then modifies a subset of the model weights to produce the desired activation for a whole batch of such edits. 

Let $h^L$ be the $L$-th activation of the last token of $s$. The target change is $z^L = h^L + \delta$ where $\delta$ is optimized to according to the following:

\begin{equation}
    \delta = \argmax_\delta \frac{1}{P}\sum_{j=1}^{P}\log \mathbf{P}_{\theta(h^L += \delta)}[o^*|x_j \oplus t_r(s)]
\end{equation}

This calculates a residual vector to be added to $h^L$ such that when $t_r(s)$ is input, the model outputs $o^*$ with maximum probability. $\theta(h^L += \delta)$ represents the LLM in which $\delta$ is added to $h^L$. A weight decay penalty is applied to $\delta$ to minimize the change. $x_j$ are various contextual inputs pre-pended to $t_r(s)$ to improve generalization. 

After computing each desired $\delta_i$ for each memory $i$ to be batch edited, MEMIT then computes how to spread those changes across a set of "critical layers" $[l_0:L$]. Each $\delta_i$ is split into a residual $r_i$ to be added to each layer according to $r_i = \frac{\delta_i}{L-l+1}$. This assigns each of the critical layers an approximately equal amount of modification towards the desired $\delta_i$. The following optimization is then applied to the output matrix of the Multi-Layer Perceptron (MLP) block for each critical layer:

\begin{equation}
    \begin{split}
        \hat{W}_l^{out} = \argmin_{\hat{W}} & \Biggl( \sum_{i=1}^n\left\|\hat{W} k_i - W_l^{out} k_i - r_i\right\|^2 \\
    & + \sum_{i=n+1}^{n+u}\left\|\hat{W} k_i - W_l^{out} k_i\right\|^2 \Biggr)
    \end{split}
\end{equation}

Where $\hat{W}_l^{out}$ is the new final weight matrix of the MLP block of layer $l$, $W_l^{out}$ is the original weight matrix, $k_i$ is the activation input to the weight matrix being changed (intermediate activation of the MLP block). This computation modifies $W_l^{out}$ so that $n$ edited associations are modified by residual $r_i$ while not affecting $u$ unchanged associations. The actual computation is done using a more efficient format that does not require recomputing large numbers of unchanged associations. Further details can be found in the original paper.

\section{Hyperparameters}

We introduce no additional hyperparameters beyond hop-count and use hyperparameters from \citet{memitMeng2022MassEditingMI} otherwise. 

%

\section{Creating New Relation Templates with LLMs}
\label{sec:llm_for_template_creation}

The composable structure of our templates for contextual edits allows us to scale K-Edit to exponentially many combinations of multi-hop relations from the number of base templates we have. We note that this approach easily meets the needs of our experiments. 

In actual deployments it may be worthwhile to include methods for scaling the number of relation types even further. Many knowledge graphs include thousands of different relation types. We introduce an LLM based approach to automatically constructing these base templates. The prompts we use can be found in figures \ref{fig:prompt_for_template1} and \ref{fig:prompt_for_template2}. 

\end{document}